\documentclass[runningheads]{llncs}

\usepackage{eccv}

\usepackage[letterpaper,hmargin=1.5in,vmargin=1in,footskip=0.5in]{geometry}

\makeatletter
\def\ps@plainpreprint{%
  \let\@mkboth\@gobbletwo
  \let\@oddhead\@empty \let\@evenhead\@empty
  \def\@oddfoot{\hfil\normalfont\thepage\hfil}%
  \def\@evenfoot{\hfil\normalfont\thepage\hfil}%
  \def\chaptermark##1{}\def\sectionmark##1{}\def\subsectionmark##1{}}
\makeatother
\usepackage{eccvabbrv}

\usepackage{graphicx}
\usepackage{booktabs}
\usepackage{longtable}  
\usepackage{array}      

\usepackage{hyperref}

\begin{document}

\title{VGIQA: Fine-Grained Image Quality Assessment with Visual Grounding Rewards} 

\titlerunning{VGIQA}
\authorrunning{W.~Wang et al.}

\author{Weichang Wang\inst{1,\ast}, Liyu Chen\inst{2,\ast}, Shuyue Guan\inst{2},
Jie Hu\inst{2}, Fei Gao\inst{1}, Wei Li\inst{2,\dagger}, Nannan Wang\inst{1,\dagger}}

\institute{State Key Laboratory of Integrated Services Networks, Xidian University \and
Huawei Foundation Model Department\\
\email{weichangwang@stu.xidian.edu.cn}, \email{wei.lee@huawei.com}}

\maketitle

\makeatletter
\def\@fnsymbol#1{\ensuremath{\ifcase#1\or\ast\or\dagger\or\ddagger\or
   \mathchar "278\or \mathchar "27B\or \|\or \ast\ast\or \dagger\dagger
   \or \ddagger\ddagger \else\@ctrerr\fi}}
\makeatother
\renewcommand{\thefootnote}{\fnsymbol{footnote}}%
\footnotetext[1]{Both authors contributed equally to this research.}%
\footnotetext[2]{Corresponding author.}%
\renewcommand{\thefootnote}{\arabic{footnote}}%

\begin{figure}[tb]
  \centering
  \includegraphics[height=7.5cm]{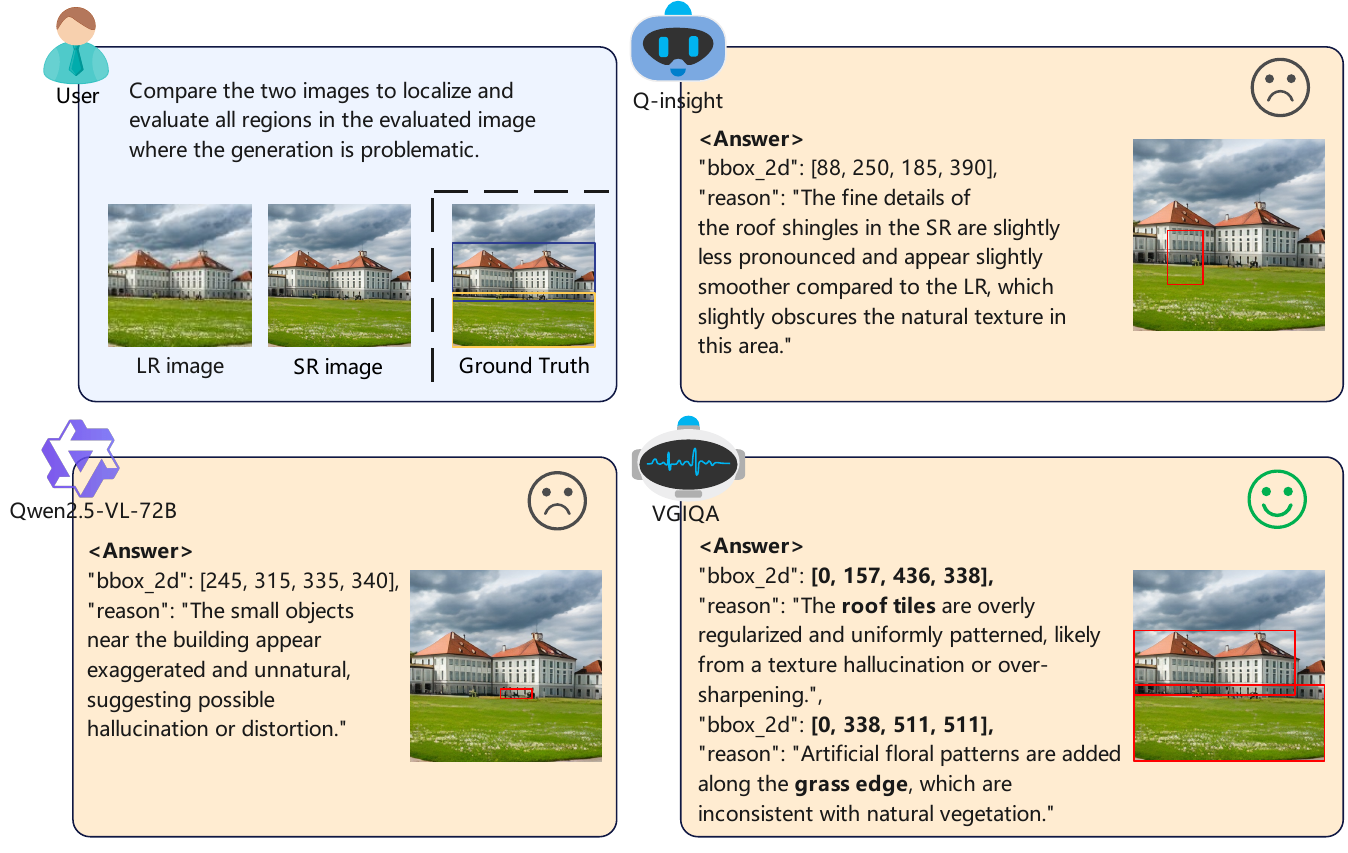}
  \caption{\textbf{Qualitative visualization of the local quality assessment results.} Given a reference image, a distorted image to be evaluated, and a user query, VGIQA produces more reliable and accurate responses compared with the state-of-the-art RL-based IQA method Q-insight and the general large vision-language model Qwen2.5-VL-72B-Instruct\cite{qwen2.5}.}
  \label{intro1}
\end{figure}

\begin{abstract}
  Image quality assessment (IQA) aims to evaluate the perceptual visual quality of images. Recently, the rapid development of Multi-modal Large Language Models (MLLMs) has significantly advanced the IQA field. By integrating chain-of-thought (CoT) reasoning and reinforcement learning (RL), these models can generate more accurate and interpretable results while relying on less training data. However, existing MLLM-based methods are predominantly constrained to holistic, image-level evaluations and lack the capability to produce fine-grained, localized assessments, which substantially limits their practical applicability. To address this limitation, we propose VGIQA, a novel IQA model endowed with both local and global quality assessment capabilities. VGIQA is trained via a two-stage paradigm comprising a cold-start phase followed by reinforcement learning. Driven by carefully designed local quality assessment reward functions, our model exhibits exceptional image quality understanding on VGSR-2K, a new dataset we propose specifically for local quality evaluation, significantly outperforming existing open-source MLLMs such as Qwen2.5-VL-72B-Instruct\cite{qwen2.5} and Qwen3-VL-32B-Instruct\cite{qwen3}. Furthermore, the local quality evaluation task can effectively enhance the model's performance on the degradation perception task.
\end{abstract}

\section{Introduction}
\label{sec:intro}

Image Quality Assessment (IQA) serves as a foundational pillar in computer vision, underpinning tasks ranging from classical restoration (e.g., super-resolution, denoising) to the surging field of AI-Generated Content (AIGC). It provides a quantitative and objective measure for evaluating the perceptual quality of visual content, which is crucial for guiding model training, optimizing generative processes, detecting artifacts, and ensuring the reliability of visual systems in practical deployment. 

Early IQA research primarily focused on handcrafted feature-based approaches, which rely on manually designed statistical or perceptual metrics to assess image quality. Representative methods include PSNR, SSIM\cite{wang2004image}. However, these handcrafted methods suffer from inherent limitations: they exhibit poor generalization across diverse distortion types, as their design is constrained by predefined perceptual or statistical assumptions rather than data-driven learning. Then, the subsequent era of Deep Learning, represented by NIMA\cite{talebi2018nima}, ManIQA\cite{yang2022maniqa} and CLIP-IQA\cite{wang2023exploring} methods, significantly improved alignment with human subjective scores by leveraging data-driven deep features. Compared to handcrafted approaches, these IQA methods achieve superior alignment with human subjective evaluations, as they can automatically learn high-level perceptual features and complex distortion patterns from large-scale datasets. They also demonstrate stronger generalization across diverse distortion types, making them more suitable for real-world vision tasks. Nevertheless, these methods still have notable drawbacks: They treat images as holistic distributions, offering a single quality score while remaining silent on where and why the quality degradation occurs. This lack of spatial interpretability is particularly problematic for modern generative models (e.g., Diffusion Models), which frequently introduce spatially inhomogeneous artifacts---such as local texture blurring or structural distortions---amidst otherwise high-fidelity regions.

In recent years, with the rapid advancement of Multimodal Large Language Models (MLLMs), MLLM-based IQA methods have emerged as a new research paradigm, enabling interpretable and semantic-aware quality assessment by fusing visual and linguistic information. These methods can be roughly divided into two categories based on their training paradigms. The first category adopts supervised fine-tuning (SFT) on IQA datasets, such as Q-Align\cite{wu2023q}, DeQA\cite{you2025teaching}, and DepictQA\cite{you2024depicting}. These methods leverage annotated quality scores or constructed text descriptions to align MLLMs with human subjective judgments, achieving impressive performance in both global quality scoring and descriptive quality evaluation. However, they heavily rely on large-scale, high-quality annotated IQA data, which limits their scalability and generalization to unseen distortion types. The second category employs reinforcement learning (RL) to align with human preferences, such as Q-insight\cite{li2025q}. By designing reward functions that reflect human perceptual preferences, these methods eliminate the reliance on large-scale annotated data, enhance the model's adaptability to diverse quality scenarios, and better align with human subjective judgments. Critically, modern low-level vision and generative models frequently suffer from spatially inhomogeneous quality degradation---such as local artifacts, spurious textures, distorted details, and structural anomalies---due to their optimization objectives based on global losses (prioritizing global statistical rationality) and the lack of local regularity constraints. Yet, all existing MLLM-based IQA methods lack the capability of fine-grained local quality assessment, failing to explicitly locate and evaluate these local quality defects. Even some general large-scale MLLMs perform poorly on this task, as shown in Figure \ref{intro1}. Therefore, an IQA model capable of accurately perceiving, localizing, and suppressing local quality degradation is crucial for enhancing the reliability and realism of modern low-level vision and generative models.

To bridge this gap, we propose VGIQA, the MLLM-based framework that integrates visual grounding with fine-grained IQA. Our key insight is that global quality scoring should be grounded in verifiable local evidence. To achieve this, we introduce a novel training paradigm combining a "Cold Start" phase with Group Relative Policy Optimization (GRPO). Unlike traditional RLHF, GRPO enables the model to engage in a self-reflective reasoning process by comparing multiple generated outputs within a group. We elaborately design a reward system that incorporates geometric precision metrics (DIOU) and semantic alignment (F1 Score) to enforce the model's ability to "look and then judge." To support this, we construct the VGSR-2K dataset based on the public DiffIQA and SRbench datasets. This dataset adopts inference results from multiple general large models as baselines, followed by human voting and correction, forming the first local quality assessment dataset aligned with human preferences. VGSR-2K contains pairs of reference images and AI-generated images, along with corresponding text-based quality evaluations that include both global and local quality assessments---with the latter incorporating coordinate information for visual localization. Ultimately, we perform joint training on local quality assessment, degradation perception, and global score regression, yielding a unified local-and-global image quality assessment framework.

Experimental results demonstrate that, after joint training, the degradation perception task achieves significant improvements over Q-insight; meanwhile, VGIQA's local quality assessment capability outperforms not only Q-insight but also general larger-scale MLLM, such as Qwen2.5-VL-72B-Instruct\cite{qwen2.5} and Qwen3-VL-32B-Instruct\cite{qwen3}, as shown in Figure \ref{intro1}. To summarize, our contributions are three-fold: \\
(1) We propose VGIQA, the MLLM-based fine-grained visual quality assessment model. Unlike previous works that only provide global quality scores and descriptive evaluations for the entire image, our method can explicitly and directly identify local quality issues in images, even outperforming existing general-purpose VL large models.\\
(2) We design a multi-task joint training framework. Through elaborately designed reward functions, VGIQA is capable of effectively handling both global and local quality assessment tasks.\\
(3) We release the first reference-based fine-grained visual grounding IQA dataset. The VGSR-2K dataset provides both a training set and a benchmark, enabling the community to improve and evaluate the spatial interpretability of quality assessment models.
\begin{figure}[t]
  \centering
  \includegraphics[width=\linewidth]{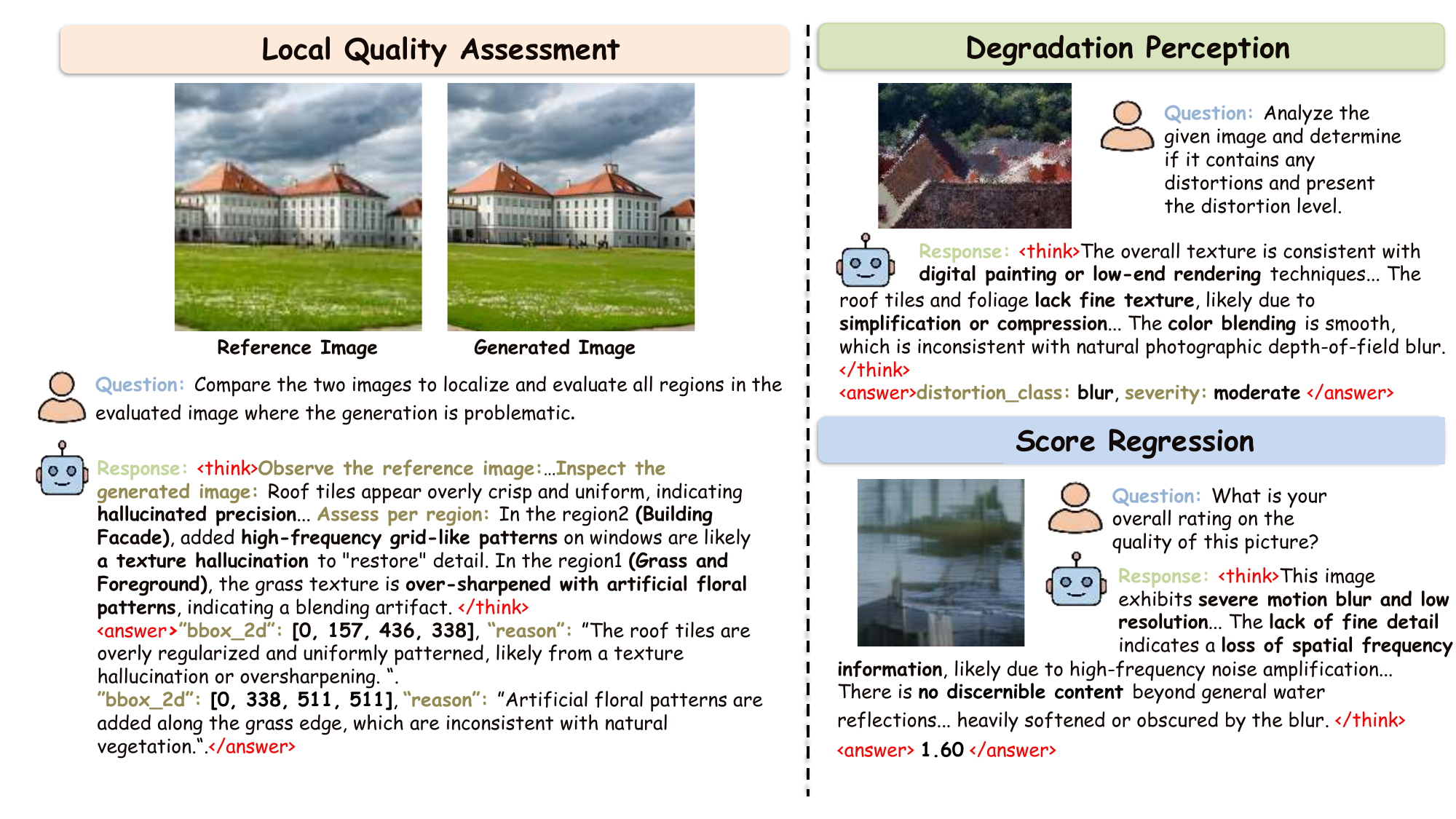}
  \caption{\textbf{Example applications of the proposed VGIQA framework.} Built upon a Qwen-based foundation model, VGIQA achieves comprehensive quality understanding via multi-task reinforcement learning. While existing models (e.g., Q-Insight) are limited to holistic \textbf{Score Regression} and \textbf{Degradation Perception}, VGIQA explicitly bridges the gap in fine-grained \textbf{Local Quality Assessment}. Driven by novel visual grounding rewards, it adopts a rigorous "look and then judge" paradigm---precisely localizing spatial artifacts before formulating final judgments.}
  \label{fig:intro_tasks}
\end{figure}

\section{Related Works}
\label{sec:Related Works}

Early image quality assessment (IQA) employs handcrafted features to score image quality and is primarily categorized into full-reference (FR-IQA) \cite{wang2004image, sheikh2006image, zhang2011fsim} and no-reference (NR-IQA) \cite{mittal2012no, saad2012blind, ma2017learning, mittal2012making, moorthy2010two, moorthy2011blind} methods. FR-IQA evaluates image quality by calculating the similarity between the distorted image and a high-quality reference image. For instance, SSIM \cite{wang2004image} computes brightness, contrast, and structural similarity separately and then performs weighted fusion; VIF \cite{sheikh2006image} calculates the proportion of useful information retained after distortion; FSIM\cite{zhang2011fsim} measures the phase (edge, contour, and texture position and direction) that the human eye is most sensitive to, which is similar to the gradient features derived from the Sobel operator. NR-IQA method relies on manually designed statistical features in the spatial or frequency domain to predict image quality without a reference image. For example, BRISQUE\cite{mittal2012no} extracts generalized Gaussian distribution (GGD) natural statistical features from locally normalized residuals in the spatial domain; BLIINDS\cite{saad2012blind} models the GGD distribution of each subband coefficient in the DCT frequency domain to capture distortions such as compression and blocking artifacts. While the above methods rely on handcrafted features for image quality scoring, they cannot accurately evaluate complex practical scenarios involving local distortions or real-world distortions.

With the advancement of neural networks, deep learning-based IQA models \cite{ke2021musiq, wang2023exploring, sun2022graphiqa, kang2014convolutional, liu2017rankiqa, pan2018blind, su2020blindly, zheng2021learning, zhu2020metaiqa, yang2022maniqa} have emerged. NIMA\cite{talebi2018nima} employs deep networks to predict the distribution of image quality scores; ManIQA\cite{yang2022maniqa} incorporates multi-scale features and attention mechanisms to enhance focus on sensitive regions; MUSIQ\cite{ke2021musiq} utilizes Transformers and multi-scale slicing to evaluate the quality of images of any size; CLIP-IQA\cite{wang2023exploring} leverages CLIP's capability to align image and text semantics, improving its generalization across different datasets and scenarios. These IQA models typically treat images as a whole, providing a single overall quality score without specifying the location or cause of quality degradation, and failing to align with human preferences.

Multimodal large language models (MLLMs) with visual understanding capabilities have demonstrated promising performance in image quality evaluation \cite{chen2024q, jia2025refine, li2025q, wu2023qbench, wu2023q, wu2024q, wu2024towards, xie2026q, you2024depicting, you2024descriptive, zhang2024q, zhang2025q, zhang2025teaching}. One approach employs supervised fine-tuning on image quality assessment datasets. For instance, Q-Instruct\cite{wu2024q} creates a multimodal image quality assessment dataset to enhance the image perception capabilities of MLLMs. Q-Align\cite{wu2023q} evaluates image quality by calculating the alignment between image features and the semantic features of "high-quality images", using pairwise ranking loss or direct regression of Mean Opinion Score (MOS). DepictQA\cite{you2024depicting} utilizes paired training data consisting of "image-quality description text" pairs, enabling MLLMs to generate natural language descriptions of distortion types, locations, severity, and their impact on visual content. Another approach leverages reinforcement learning (RL) \cite{christiano2017deep, hui2024qwen2, silver2017mastering, yang2024qwen2, ying2024internlm, zhang2024o1, sun2024aligning, yu2024rlaif, yu2024rlhf, zhao2023beyond, shao2024deepseekmath} to align with human preferences. For example, Q-Insight\cite{li2025q} designs scoring rewards, degradation classification rewards and intensity perception rewards, employing GRPO\cite{shao2024deepseekmath} reinforcement learning to improve the prediction accuracy of subjective scores, distortion types, and severity, significantly enhancing the model's cross-domain generalization and reasoning reliability. Q-Hawkeye\cite{xie2026q} introduces an uncertainty-aware reward mechanism to increase the weight of stable samples, reduce noise interference, and improve model accuracy. Refine-IQA\cite{jia2025refine} designs a probability difference reward to specifically supervise the model's "reasoning process." However, these MLLM-based image quality assessment methods lack fine-grained local quality assessment capabilities and cannot precisely locate and evaluate local quality defects.

\section{Methodology}

\subsection{Overview}
\label{subsec:overview}

\begin{figure}[t]
\centering
\includegraphics[width=\linewidth]{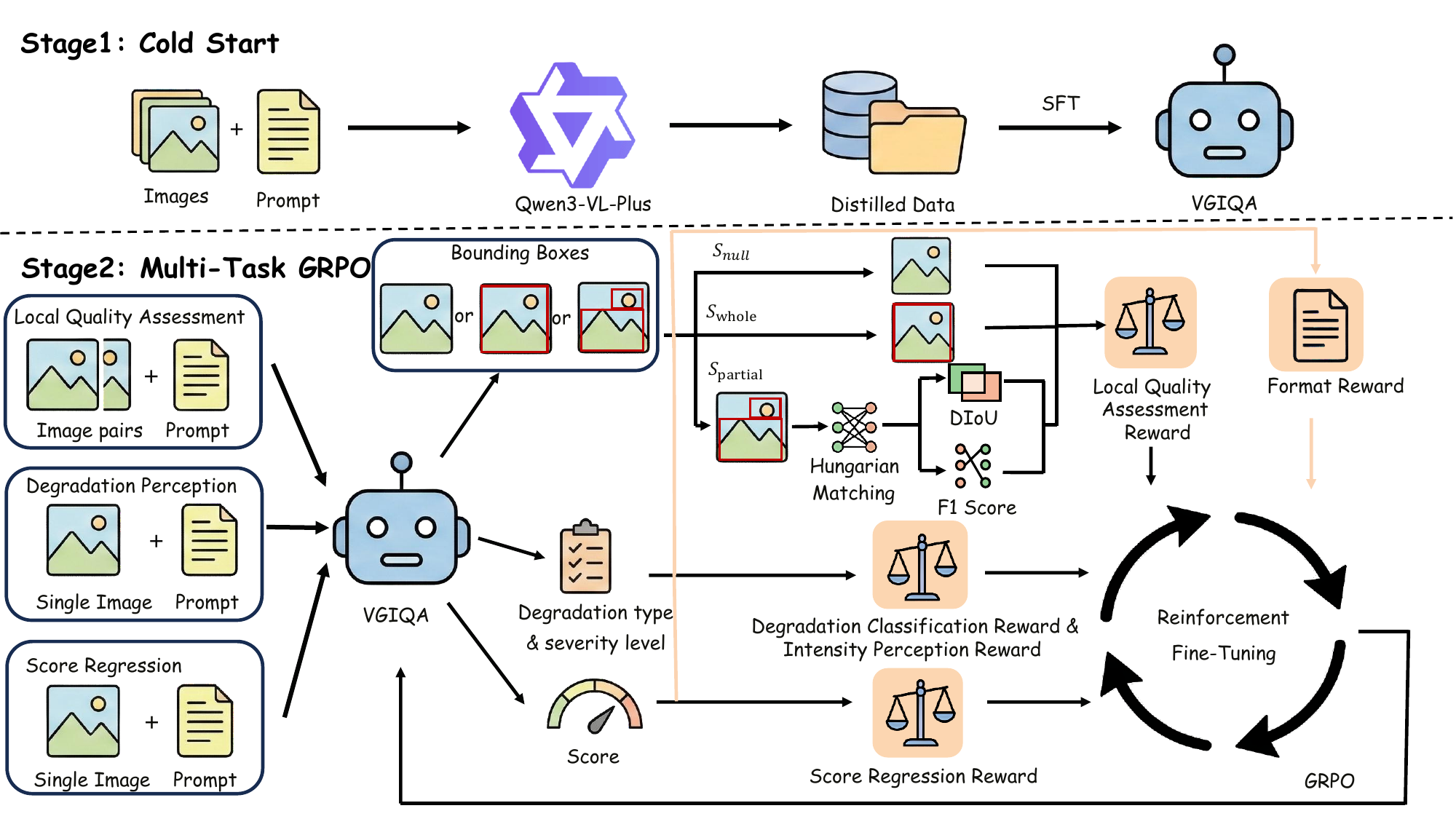}
\caption{\textbf{Overview of the proposed VGIQA framework.} The framework follows a two-stage training paradigm. \textbf{Stage 1 (Cold Start):} The Qwen-based foundation model is initialized via SFT using CoT reasoning data distilled from Qwen3-VL-Plus\cite{qwen3} to establish structural formatting and inject IQA priors. \textbf{Stage 2 (Multi-Task GRPO):} Reinforcement learning jointly optimizes Local Quality Assessment, Degradation Perception, and Score Regression. Alongside a shared format reward, task-specific rewards are dynamically routed: a dedicated local reward exclusively supervises the local task (computing zero for non-local tasks), evaluating spatial bounding boxes via Hungarian Matching, negative DIoU, and F1 Score. Meanwhile, global tasks are optimized via verifiable score and classification/intensity rewards, unifying local-and-global quality evaluation.}
\label{fig:overview}
\end{figure}

To address the inherent limitations of existing MLLMs that predominantly rely on holistic evaluations, we propose VGIQA, a unified framework explicitly endowed with both fine-grained local and global quality perception capabilities. As illustrated in Fig.~\ref{fig:overview}, VGIQA is driven by a two-stage training paradigm. First, a \textit{Cold Start} phase aligns the reasoning format and injects domain-specific IQA priors via supervised fine-tuning. Subsequently, a Multi-Task GRPO stage jointly optimizes local and global assessments to align with human preferences. A core algorithmic breakthrough of our framework is the strict reward isolation mechanism: the local visual grounding reward is exclusively routed to the local quality assessment task. This ensures that the precise spatial localization of defective regions is strictly decoupled from holistic global tasks (\ie, degradation perception and score regression), preventing objective confusion and inter-task interference.

\subsection{Cold Start: Initializing IQA-Specific CoT Reasoning}
\label{subsec:cold_start}

Directly applying large-scale reinforcement learning to MLLMs without proper initialization often leads to unconstrained exploration and unstable convergence. To mitigate this, we introduce a \textit{Cold Start} phase to explicitly align the model's reasoning trajectories with structured formats and inject domain-specific IQA knowledge. To guarantee broad generalizability across diverse visual generation and restoration scenarios, we formalize the visual inputs as a unified image group comprising a reference image ($I_{ref}$) and an evaluated image ($I_{eval}$). 

Specifically, we curate a high-quality Chain-of-Thought (CoT) dataset, denoted as $\mathcal{D}_{CS}$, comprising $2,288$ structurally rigorous samples meticulously distilled from the Qwen3-VL-Plus\cite{qwen3} model. For each training instance $(x, y) \in \mathcal{D}_{CS}$, the input $x$ encompasses $\{I_{ref}, I_{eval}\}$ coupled with a task-specific textual prompt, while $y = \{y_1, y_2, \dots, y_T\}$ represents the target structured CoT response sequence of length $T$. The Qwen-based foundation model is then initialized via Supervised Fine-Tuning (SFT) using the standard autoregressive negative log-likelihood loss:
\begin{equation}
\mathcal{L}_{CS}(\theta) = - \sum_{i=1}^{T} \log p(y_i | x, y_{<i}; \theta),
\end{equation}
where $\theta$ denotes the trainable parameters. This optimization objective explicitly endows the model with highly interpretable, step-by-step reasoning capabilities and professional terminology, establishing a robust policy initialization for the subsequent multi-task GRPO phase.
\subsection{Multi-Task Visual Reinforcement Fine-Tuning via GRPO}
\label{subsec:grpo}

\subsubsection{Rewards for Local Quality Assessment}
\label{subsubsec:local_reward}

To endow the model with precise spatial localization capabilities, we evaluate predicted bounding boxes $\mathcal{B}^{p} = \{b^p_1, \dots, b^p_m\}$ against ground truths $\mathcal{B}^{g} = \{b^g_1, \dots, b^g_n\}$. We first categorize the predictions into three macro-scene types $c \in \{\mathcal{S}_{empty}, \mathcal{S}_{whole}, \mathcal{S}_{partial}\}$. A class-balanced base reward $r_{base} = 10 / p_c$ is assigned for correctly predicting $c$, where $p_c$ is the class frequency, mitigating spatial imbalance. For the non-trivial $\mathcal{S}_{partial}$ scenario, we solve an optimal bipartite matching problem via the Hungarian algorithm using negative Distance Intersection over Union (DIoU) as the cost matrix $\mathbf{C} \in \mathbb{R}^{m \times n}$. To provide continuous gradient signals for coordinate regression, we define $r_{cont}$ as the average normalized DIoU of the $|\pi|$ matched pairs:
\begin{equation}
r_{cont} = \frac{1}{|\pi|} \sum_{i=1}^{|\pi|} \frac{1 + \text{DIoU}(b^p_i, b^g_{\pi(i)})}{2}.
\end{equation}
The overall local quality assessment reward $r_{local}$ is formulated as:
\begin{equation}
r_{local} = r_{base} + \omega_{local} (F_1 + r_{cont}),
\end{equation}
where $\omega_{local}$ is a dynamic weighting coefficient preventing the model from collapsing into trivial solutions (\eg, predicting only empty boxes) during early training.

\subsubsection{Overall Multi-Task Reward Formulation}
\label{subsubsec:other_rewards}

To seamlessly unify local localization with holistic quality perception, we integrate our local reward with global tasks strictly following the configurations of Q-Insight~\cite{li2025q}. Specifically, we define a structural format reward $r_{format}$, a binary score regression reward $r_{score}$ (activated if the predicted score falls within an $\epsilon$-threshold of the ground-truth MOS), and two degradation perception rewards: $r_{deg}$ for correct distortion type classification, and $r_{lev}$ for precise severity level matching.

Crucially, to implement the strict reward isolation mechanism highlighted in \cref{subsec:overview}, the overall multi-task reward $r_{total}$ dynamically routes task-specific gradients using indicator functions:
\begin{equation}
r = r_{format} + \lambda_{local} r_{local} + \lambda_{score} r_{score} + \lambda_{deg} (\alpha_{1} r_{deg} + \alpha_{2} r_{lev}).
\end{equation}
Here, $\lambda_{local}$, $\lambda_{score}$, and $\lambda_{deg}$ serve as task-routing gating variables---evaluating to 1 exclusively for the currently sampled task and 0 otherwise. This mathematical formulation explicitly decouples local spatial reasoning from global holistic evaluation, effectively preventing inter-task interference during the GRPO phase.

\section{Proposed Dataset: VGSR-2K}
\label{sec:dataset}

\subsection{Overview of VGSR-2K}
\label{subsec:dataset_overview}

Despite the rapid advancements in MLLMs for IQA, the research community currently lacks high-quality datasets exclusively tailored for fine-grained, local quality assessment. Existing datasets predominantly provide holistic, image-level scores or textual descriptions, which are insufficient to train models for the precise spatial localization of local erroneous regions. To bridge this critical gap, we introduce VGSR-2K, a novel visual grounding dataset strictly constructed for fine-grained, local quality assessment. Built upon the foundational data from DiffIQA \cite{diffiqa} and SRBench \cite{diffiqa}, the data instances in VGSR-2K are organized as paired image groups. Specifically, each sample typically comprises a Low-Resolution (LR) reference image coupled with either a Super-Resolved (SR) generated image or a High-Resolution (HR) ground-truth image (i.e., $[I_{LR}, I_{SR}]$ or $[I_{LR}, I_{HR}]$ pairs). The ground truth of this dataset is provided in the form of spatial bounding boxes, precisely grounding the specific locations of the erroneous regions within the evaluated images.

\subsection{Pipeline}
\label{subsec:annotation_pipeline}

\begin{figure}[t]
  \centering
  \includegraphics[width=\linewidth]{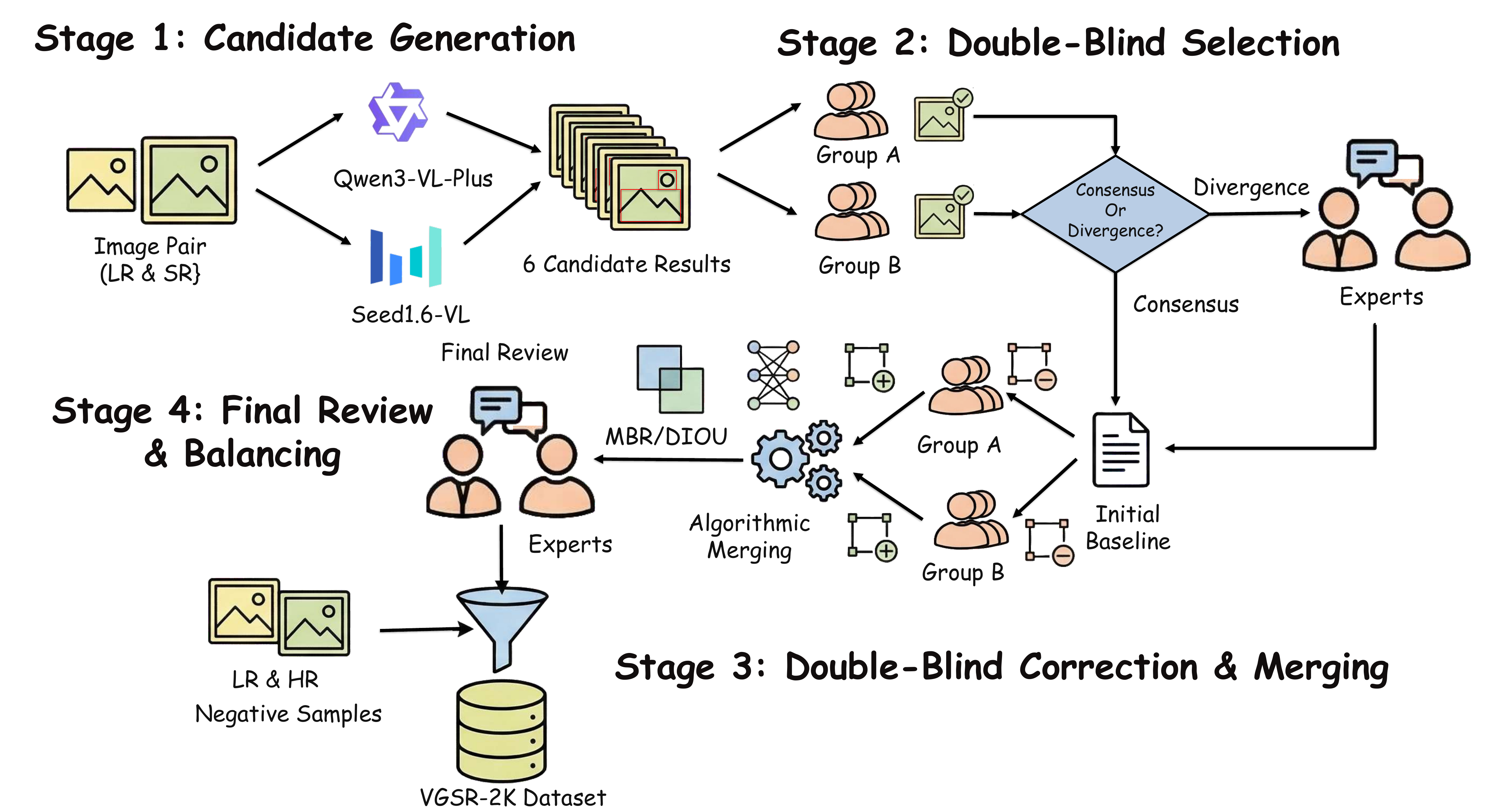}
  \caption{\textbf{Overview of the VGSR-2K dataset construction pipeline.} The dataset is curated via a rigorous human-in-the-loop pipeline. \textbf{(1) Candidate Generation:} Qwen3-VL-Plus\cite{qwen3} and Seed1.6-VL independently process LR-SR image pairs, yielding six candidate spatial bounding boxes. \textbf{(2) Double-Blind Selection:} Two annotator groups select the most consistent candidate; consensus establishes the baseline, while divergence prompts expert resolution. \textbf{(3) Double-Blind Correction \& Merging:} Annotators refine baselines by adding or removing boxes. Corrections are merged via the Hungarian algorithm with a negative DIoU cost metric, extracting the MBR for matched pairs. \textbf{(4) Final Review \& Balancing:} After expert review, pristine LR-HR image pairs are introduced as negative samples (with empty annotations) to calibrate spatial predictions and prevent trivial over-generation.}
  \label{fig:dataset_pipeline}
\end{figure}

As illustrated in Fig.~\ref{fig:dataset_pipeline}, we design a rigorous human-in-the-loop pipeline to ensure the precision and consistency of the spatial annotations. Based on a pilot study, we establish a protocol to explicitly annotate three categories of erroneous regions: (a) \textit{Generation Artifacts} (\eg, structural distortions or visual noise); (b) \textit{Semantic Inconsistency} (\ie, semantic deviations from the LR reference); and (c) \textit{Unrealistic Hallucinations} (\ie, spurious contents violating physical laws). The pipeline proceeds in four stages.
\subsubsection{Candidate Generation.} We utilize two state-of-the-art MLLMs, Qwen3-VL-Plus\cite{qwen3} and Seed1.6-VL, for initial proposal generation. For each of the $1,798$ $[I_{LR}, I_{SR}]$ image pairs, each model performs three independent inferences, yielding six candidate spatial bounding box sets per pair.
\subsubsection{Double-Blind Selection.} Four experienced annotators are equally divided into two independent groups (Group A and Group B). They contrast the LR and SR images to blindly select the most visually consistent candidate among the six. Consensual selections are directly adopted as initial baselines. For divergent samples, two senior IQA experts intervene to determine the optimal baseline.
\subsubsection{Double-Blind Correction and Merging.} To prevent manual coordinate-level bias, Groups A and B independently refine the baselines by strictly deleting incorrect boxes or supplementing missing ones, without fine-tuning existing coordinates. These independent corrections are algorithmically merged: we take the intersection of the retained box indices from both groups. For supplemented boxes, we formulate a bipartite matching problem via the Hungarian algorithm with negative DIoU as the cost matrix. Matched box pairs with $\text{DIoU} > 0.5$ are preserved, and their Maximum Bounding Rectangle (MBR) is computed as the final merged annotation.
\subsubsection{Final Review \& Balancing.} Finally, two senior experts conduct a comprehensive review of the algorithmically merged samples to resolve any residual ambiguities, ensuring high-fidelity spatial ground truths.

\subsection{Dataset Statistics and Distribution}
\label{subsec:statistics}

Following the rigorous annotation pipeline, the $1,798$ meticulously verified data groups, denoted as $[I_{LR}, I_{SR}, \text{result}_x]$, are partitioned into identically distributed training and testing sets. A critical challenge in visual grounding for quality assessment is the potential for trivial over-generation, driven by the severe spatial imbalance between explicitly annotated localized artifacts and vast background areas. To counteract this, we purposefully introduce pristine $[I_{LR}, I_{HR}]$ pairs into both sets. Inherently lacking any defective areas, these reference pairs carry empty spatial annotations, serving as high-quality negative samples to systematically calibrate the model's spatial prediction behavior.

The dataset is systematically constructed from two distinct subsets to encompass diverse generative flaws. For the DiffIQA subset, we allocate $698$ $[I_{LR}, I_{SR}]$ groups for training and $100$ for testing. To maintain balance, we supplement these with $102$ and $18$ negative $[I_{LR}, I_{HR}]$ pairs, yielding $800$ training and $118$ testing samples, respectively. The training sources are rigorously balanced across diverse domains, including DF2K ($264$), the Internet ($266$), and camera captures ($270$), with progressive low-resolution degradation intensities (Levels 1 to 6) proportionately distributed as $133$, $127$, $135$, $149$, $147$, and $109$. The testing set maintains this identical distribution, comprising domains of DF2K ($32$), the Internet ($33$), and camera captures ($53$), alongside corresponding degradation levels of $22$, $18$, $18$, $21$, $25$, and $14$.

To further enhance robustness against unknown distortions, the SRBench subset incorporates multi-paradigm data spanning $100$ natural scenes. Each scene intrinsically provides one HR reference and $10$ corresponding SR images generated by diverse architectures (\ie, regression-based, GAN-based, and diffusion-based models). We randomly allocate $80$ scenes to formulate the training set, yielding $800$ $[I_{LR}, I_{SR}]$ groups and $80$ $[I_{LR}, I_{HR}]$ negative pairs (totaling $880$ samples). The remaining $20$ scenes are exclusively allocated to the testing set, encompassing $200$ positive and $20$ negative pairs (totaling $220$ samples). This highly stratified organization ensures that VGSR-2K comprehensively encapsulates a broad spectrum of visual characteristics and spatial distributions of flawed regions, providing a statistically robust foundation for fine-grained quality assessment.

\section{Experiments}

\begin{table}[t]
    \centering
    \begin{minipage}[t]{0.48\linewidth}
        \centering
        \caption{\textbf{Quantitative comparison of the local quality assessment task} on the VGSR-2K dataset. Qwen2.5-VL-72B and Qwen3-VL-32B denote Qwen2.5-VL-72B-Instruct and Qwen3-VL-32B-Instruct, respectively. VGIQA significantly outperforms both large-scale general MLLMs and the IQA-specific baseline. The best and second-best results are highlighted in \textbf{bold} and \underline{underlined}, respectively.}
        \label{tab:local_qa}
        \resizebox{\linewidth}{!}{
            \begin{tabular}{>{\raggedright\arraybackslash}p{0.55\linewidth}cc}
                \toprule
                Method & F1 Score & DIoU \\
                \midrule
                Q-Insight \cite{li2025q} & 0.1452 & 0.1673 \\
                Qwen2.5-VL-72B\cite{qwen2.5} & \underline{0.1770} & 0.1997 \\
                Qwen3-VL-32B\cite{qwen3} & 0.1756 & \underline{0.2185} \\
                VGIQA (Ours) & \textbf{0.3024} & \textbf{0.3235} \\
                \bottomrule
            \end{tabular}
        }
    \end{minipage}\hfill
    \begin{minipage}[t]{0.48\linewidth}
        \centering
        \caption{\textbf{Ablation study of each training stage} on the local quality assessment task. The base model is Qwen2.5-VL-7B-Instruct\cite{qwen2.5}. The GRPO phase is crucial for the improvement in Local Quality Assessment. The best and second-best results are highlighted in \textbf{bold} and \underline{underlined}, respectively.}
        \label{tab:ablation_stage}
        \vspace*{22pt}
        \resizebox{\linewidth}{!}{
            \begin{tabular}{>{\raggedright\arraybackslash}p{0.55\linewidth}cc}
                \toprule
                Method & F1 Score & DIoU \\
                \midrule
                Base Model & \underline{0.1497} & 0.1672 \\
                Base Model + Cold Start & 0.1455 & \underline{0.2013} \\
                Base Model + Cold Start + GRPO (VGIQA) & \textbf{0.3024} & \textbf{0.3235} \\
                \bottomrule
            \end{tabular}
        }
    \end{minipage}
\end{table}
\subsection{Experimental Setup}

\subsubsection{Datasets.} For the local quality assessment task, we train and evaluate VGIQA using the proposed VGSR-2K dataset (detailed in \cref{subsec:statistics}). During training, the diverse data subsets are thoroughly mixed, whereas evaluations are conducted separately on their respective testing subsets. For global assessment tasks, we strictly follow the configurations of Q-Insight \cite{li2025q}. Specifically, the score regression task utilizes identical splits across established benchmarks (\eg, KonIQ, SPAQ, KADID, PIPAL, LIVE-Wild, AGIQA-3K, and CSIQ). For the degradation perception task, we adopt the identical training set from DQ-495K and randomly sample $1,000$ identically distributed images to formulate our test set.

\subsubsection{Evaluation Metrics.} 
We adopt distinct assessment metrics tailored to the specific characteristics of each task. For the local quality assessment task, we utilize the F1 Score and DIoU, as formulated in \cref{subsubsec:local_reward}, to rigorously measure the spatial localization accuracy of erroneous regions. For the score regression task, we report the Pearson Linear Correlation Coefficient (PLCC) and Spearman Rank-order Correlation Coefficient (SRCC) to evaluate the consistency between model predictions and human subjective scores. For the degradation perception task, performance is measured using the prediction accuracy of degradation categories (Deg.Acc.) and degradation levels (Lev.Acc.).

\subsubsection{Implementation Details.} 
We adopt Qwen-2.5-VL-7B-Instruct\cite{qwen2.5} as our base model. The training encompasses a 3-epoch Cold Start phase and a 10-epoch GRPO phase. Elaborate hyperparameter settings and training configurations are detailed in \cref{subsec:implementation_details} of the Appendix.

\subsection{Local Quality Assessment}
We first evaluate VGIQA on the local quality assessment task, which aims to precisely localize erroneous regions in evaluated images. As this is a newly introduced fine-grained assessment paradigm, existing MLLM-based IQA models inherently lack the capability to directly output explicit spatial coordinates for erroneous regions. To demonstrate the superiority of our approach, we compare VGIQA with current state-of-the-art large-scale general-purpose MLLMs, including Qwen2.5-VL-72B-Instruct\cite{qwen2.5} and Qwen3-VL-32B-Instruct\cite{qwen3}, as well as an IQA-specific SOTA model of the same parameter scale, Q-Insight \cite{li2025q}.
As shown in \cref{tab:local_qa}, VGIQA achieves substantial performance gains on the VGSR-2K dataset. Remarkably, even when compared to general-purpose MLLMs with 10 times more parameters (\eg, Qwen2.5-VL-72B-Instruct\cite{qwen2.5}), VGIQA still demonstrates a commanding lead. Specifically, VGIQA attains an F1 Score of 0.3024 and a DIoU of 0.3235, far exceeding the baseline models. These results strongly validate that through our proposed two-stage training paradigm and carefully designed local rewards, VGIQA effectively fills the blank in fine-grained spatial localization for image quality assessment.

\subsection{Global Quality Assessment}
We further evaluate VGIQA on two global quality assessment tasks: degradation perception and score regression.

\subsubsection{Degradation Perception}
We evaluate the capability of VGIQA in identifying global physical degradations (\eg, Blur, JPEG Compression, Darken, and Noise) and predicting their severity levels. As shown in \cref{tab:deg_perc}, VGIQA demonstrates a clear advantage over the baseline Q-Insight \cite{li2025q}. Benefiting from the multi-task joint training framework, focusing on local erroneous regions encourages the model to acquire a keener ability to capture global features. Consequently, this fine-grained attention acts as a strong regularization, significantly improving the overall accuracy of the degradation perception task. Except for a slight drop in the Lev.Acc. of JPEG Compression, VGIQA achieves substantial improvements across all other categories in both Deg.Acc. and Lev.Acc. compared to Q-Insight, bringing a noticeable boost to the global image quality assessment.

\begin{table}[t]
\centering
\caption{\textbf{Quantitative comparison of the Degradation Perception task.} VGIQA significantly improves the perception accuracy of global physical degradations, especially in Noise and Null(no degradation). The best and second-best results are highlighted in \textbf{bold} and \underline{underlined}, respectively.}
\label{tab:deg_perc}
\begin{tabular}{llcccccc}
\toprule
Method & Metrics & Blur & JPEG & Darken & Noise & Null & Average \\
\midrule
{Q-Insight \cite{li2025q}}
& Deg.Acc. & \underline{91.48\%} & \underline{98.02\%} & \underline{63.23\%} & \underline{16.13\%} & \underline{64.93\%} & \underline{66.57\%} \\
& Lev.Acc. & \underline{43.05\%} & \textbf{46.04\%} & \underline{22.70\%} & \underline{6.91\%} & - & \underline{34.23\%} \\
\midrule
{VGIQA (Ours)}
& Deg.Acc. & \textbf{94.17\%} & \textbf{98.51\%} & \textbf{67.71\%} & \textbf{62.21\%} & \textbf{94.03\%} & \textbf{82.18\%} \\
& Lev.Acc. & \textbf{52.91\%} &\underline{44.55\%} & \textbf{23.77\%} & \textbf{23.96\%} & - & \textbf{43.94\%} \\
\bottomrule
\end{tabular}
\end{table}

\subsubsection{Score Regression}
In the score regression task, we evaluate the continuous regression capability of VGIQA for global image quality scoring. The results are presented in \cref{tab:score_reg}. While VGIQA achieves slightly lower PLCC and SRCC compared to Q-Insight \cite{li2025q}, it maintains highly competitive performance, being fundamentally on par with state-of-the-art methods such as DeQA \cite{deqa}. 

Mechanistically, this slight fluctuation in linear correlation is a natural manifestation of the model acquiring finer-grained assessment capabilities. By introducing sensitivity to local erroneous regions and subtle generative flaws during joint training, VGIQA inherently adopts much stricter criteria when examining images. This heightened sensitivity to microscopic imperfections leads to overall predicted scores being lower or more fluctuating, which objectively reduces the linear correlation with MOS. Therefore, this is not a degradation of capability, but rather a direct reflection that VGIQA evaluates images from a more comprehensive and fine-grained perspective.

\begin{table}[t]
\centering
\caption{\textbf{Quantitative comparison of the score regression task} between our VGIQA and other competitive IQA methods. Results are reported in the format of PLCC/SRCC. All methods except handcrafted ones are trained on the KonIQ dataset. The best and second-best results are highlighted in \textbf{bold} and \underline{underlined}, respectively.}
\label{tab:score_reg}
\resizebox{\linewidth}{!}{
\begin{tabular}{lcccccccc}
\toprule
Method & KonIQ & SPAQ & KADID & PIPAL & LIVE-Wild & AGIQA-3K & CSIQ & Average \\
\midrule
NIQE \cite{niqa} & 0.533/0.530 & 0.679/0.664 & 0.468/0.405 & 0.195/0.161 & 0.493/0.449 & 0.560/0.533 & 0.718/0.628 & 0.521/0.481 \\
BRISQUE \cite{bri} & 0.225/0.226 & 0.490/0.406 & 0.429/0.356 & 0.267/0.232 & 0.361/0.313 & 0.541/0.497 & 0.740/0.556 & 0.436/0.369 \\
NIMA \cite{nima} & 0.896/0.859 & 0.838/0.856 & 0.532/0.535 & 0.390/0.399 & 0.814/0.771 & 0.715/0.654 & 0.695/0.649 & 0.697/0.675 \\
MUSIQ \cite{ke2021musiq} & 0.924/0.929 & 0.868/0.863 & 0.575/0.556 & 0.431/0.431 & 0.789/0.830 & 0.722/0.630 & 0.771/0.710 & 0.726/0.707 \\
CLIP-IQA+ \cite{wang2023exploring} & 0.909/0.895 & 0.866/0.864 & 0.653/0.654 & 0.427/0.419 & 0.832/0.805 & 0.736/0.685 & 0.772/0.719 & 0.742/0.720 \\
ManIQA \cite{yang2022maniqa} & 0.849/0.834 & 0.768/0.758 & 0.499/0.465 & 0.457/0.452 & 0.849/0.832 & 0.723/0.636 & 0.623/0.627 & 0.681/0.658 \\
Compare2Score \cite{Compare2Score} & 0.923/0.910 & 0.867/0.860 & 0.500/0.453 & 0.354/0.342 & 0.786/0.772 & 0.777/0.671 & 0.735/0.705 & 0.706/0.673 \\
Qwen-SFT\cite{qwen3}  & 0.889/0.866 & 0.874/0.875 & 0.668/0.663 & \textbf{0.473}/0.442 & 0.734/0.728 & \underline{0.813}/0.739 & 0.674/0.650 & 0.732/0.709 \\
Q-Align \cite{wu2023q} & \underline{0.941}/\underline{0.940} & 0.886/0.887 & 0.674/0.684 & 0.403/0.419 & 0.853/\underline{0.860} & 0.772/0.735 & 0.671/0.737 & 0.705/0.752 \\
DeQA \cite{deqa} & \textbf{0.953}/\textbf{0.941} & \textbf{0.895}/\textbf{0.896} & 0.694/0.687 & \underline{0.472}/\textbf{0.478} & \textbf{0.892}/\textbf{0.879} & 0.809/0.729 & 0.787/0.744 & 0.786/\underline{0.765} \\
Q-Insight \cite{li2025q} & 0.924/0.906 & \underline{0.894}/\underline{0.891} & \textbf{0.767}/\textbf{0.764} & 0.458/\underline{0.453} & 0.881/0.850 & \textbf{0.823}/\textbf{0.764} & \textbf{0.875}/\textbf{0.817} & \textbf{0.803}/\textbf{0.780} \\
VGIQA (Ours) & 0.939/0.924 & 0.875/0.873 & \underline{0.730}/\underline{0.718} & 0.457/0.425 & \underline{0.884}/0.856 & 0.791/\underline{0.745} & \underline{0.860}/\underline{0.796} & \underline{0.791}/0.762 \\
\bottomrule
\end{tabular}
}
\end{table}
\subsection{Ablation Studies}
We conduct ablation studies to decompose the core modules of VGIQA, specifically investigating the effectiveness of the proposed two-stage training paradigm and the benefits derived from multi-task joint training.
\subsubsection{Ablation of Each Training Stage}
To evaluate the specific contributions of the Cold Start and GRPO phases to the Local Quality Assessment capability, we compare the variants across different training stages in \cref{tab:ablation_stage}. As shown, solely relying on the Cold Start phase to inject the reasoning paradigm fails to bring a substantial leap in spatial localization. The performance on F1 Score and DIoU shows merely marginal fluctuations compared to the base model. This observation verifies that while the Cold Start provides essential format alignment and domain priors, it is insufficient for precise visual grounding. Only by applying the GRPO phase, systematically guided by our carefully designed reward functions, can the model truly excite and align its fine-grained perception of erroneous regions, which effectively enhances the capability on the VGSR-2K dataset.

\section{Conclusion}
\label{sec:conclusion}
\enlargethispage{2\baselineskip}

In this paper, we address the limitation of existing MLLM-based IQA models, which are confined to holistic, image-level evaluation and lack the capability to produce fine-grained, localized assessments. To bridge this gap, we propose VGIQA, a unified framework with both local and global quality assessment capabilities. Specifically, we introduce a novel local quality evaluation task that grounds erroneous regions with bounding boxes against a pristine reference, and construct VGSR-2K, a visual grounding dataset tailored to this task, together with carefully designed reward functions. VGIQA is trained with a two-stage paradigm---a cold-start phase followed by multi-task GRPO---that jointly optimizes local and global assessment. Extensive experiments show that, while remaining competitive on score regression, VGIQA achieves two breakthroughs: (1) exceptional local quality understanding, substantially outperforming general-purpose MLLMs with ten times the parameter scale (\eg, Qwen2.5-VL-72B-Instruct); and (2) the local quality assessment effectively boosts the model's performance on the degradation perception task. We plan to extend this paradigm to video quality assessment and to the construction of reward models for tasks such as text-to-image generation and image editing.

\clearpage
\bibliographystyle{splncs04}
\bibliography{main}

\clearpage
\renewcommand{\thesection}{\Alph{section}}
\renewcommand{\thefigure}{\Alph{figure}}
\renewcommand{\thetable}{\Alph{table}}
\renewcommand{\theequation}{\Alph{equation}}
\setcounter{section}{0}
\setcounter{figure}{0}
\setcounter{table}{0}
\setcounter{equation}{0}

\section{Ablation of Multi-Task Training}
To explore the interactions among different tasks, we compare our multi-task framework with a variant trained exclusively on the global evaluation tasks. As reported in \cref{tab:ablation_multitask}, introducing the fine-grained local quality assessment task serves as a powerful spatial feature prompt. By compelling the model to meticulously localize microscopic erroneous regions, it effectively stimulates the model's potential in capturing global physical degradation clues. Consequently, the joint training scheme successfully achieves steady performance improvements in the degradation perception task, demonstrating higher accuracy in both Deg.Acc. and Lev.Acc. across almost all specific degradation categories compared to the Global-only baseline.

\begin{table}[t]
    \centering
    \caption{\textbf{Ablation study of multi-task training} on the degradation perception task. The Global tasks include score regression and degradation perception. Joint training with the local quality assessment task significantly improves global degradation perception. The best and second-best results are highlighted in \textbf{bold} and \underline{underlined}, respectively.}
    \label{tab:ablation_multitask}
    \resizebox{\linewidth}{!}{
        \begin{tabular}{llcccccc}
            \toprule
            Method & Metrics & Blur & JPEG & Darken & Noise & null & Average \\
            \midrule
            Global-only & Deg.Acc. & \underline{51.12\%} & \underline{97.52\%} & \underline{43.05\%} & \underline{43.78\%} & \underline{72.39\%} & \underline{59.96\%} \\
            & Lev.Acc. & \underline{25.11\%} & \textbf{45.05\%} & \underline{14.08\%} & \textbf{27.19\%} & - & \underline{33.63\%} \\
            \midrule
            Global + Local (Ours) & Deg.Acc. & \textbf{94.17\%} & \textbf{98.51\%} & \textbf{67.71\%} & \textbf{62.21\%} & \textbf{94.03\%} & \textbf{82.18\%} \\
            & Lev.Acc. & \textbf{52.91\%} & \underline{44.55\%} & \textbf{23.77\%} & \underline{23.96\%} & - & \textbf{43.94\%} \\
            \bottomrule
        \end{tabular}
    }
\end{table}

\section{More Methodological Details}
\label{sec:method_details}

Due to space constraints in the main paper, this section provides comprehensive details regarding the theoretical formulation of the reinforcement learning framework and the specific mathematical designs of our multi-task reward functions.

\subsection{Preliminaries of Group Relative Policy Optimization}
\label{subsec:grpo_preliminaries}

To effectively elicit the intrinsic reasoning capabilities of Multi-modal Large Language Models (MLLMs) without relying on extensive supervised fine-tuning or training an extra value model, our framework adopts Group Relative Policy Optimization (GRPO)~\cite{shao2024deepseekmath}. For a given multi-modal query $q$, GRPO samples a group of $N$ distinct candidate responses $\{o^{(1)}, o^{(2)}, \dots, o^{(N)}\}$ from the current policy model $\pi_{\theta_{old}}$. Each response $o^{(i)}$ is evaluated using task-specific rule-based reward functions to obtain a corresponding reward score $r^{(i)}$.

Instead of relying on a separate critic model to estimate the baseline, GRPO computes the relative advantage $\hat{A}^{(i)}$ for each response by normalizing the rewards within the sampled group:
\begin{equation}
    \hat{A}^{(i)} = \frac{r^{(i)} - \text{mean}(\{r^{(1)}, \dots, r^{(N)}\})}{\text{std}(\{r^{(1)}, \dots, r^{(N)}\})}.
\end{equation}
The policy model $\pi_{\theta}$ is then optimized by maximizing the objective function $J(\theta)$, which incorporates a clipped surrogate objective and a Kullback-Leibler (KL) divergence penalty to ensure stable training:
\begin{equation}
    J(\theta) = \mathbb{E}_{q \sim \mathcal{Q}, o^{(i)} \sim \pi_{\theta_{old}}} \left[ \min \left( \rho^{(i)} \hat{A}^{(i)}, \text{clip}(\rho^{(i)}, 1-\delta, 1+\delta) \hat{A}^{(i)} \right) - \beta \mathbb{D}_{KL}(\pi_{\theta} \| \pi_{\text{ref}}) \right],
\end{equation}
where $\rho^{(i)} = \frac{\pi_{\theta}(o^{(i)}|q)}{\pi_{\theta_{old}}(o^{(i)}|q)}$ denotes the probability ratio, $\delta$ is the clipping threshold, $\beta$ controls the strength of the KL penalty, and $\pi_{\text{ref}}$ is the frozen reference model.

\subsection{More Details of Rewards Design}
\label{subsec:reward_design}

To unify holistic quality perception with fine-grained spatial localization, we design a multi-task reward system. The overall reward integrates global task constraints with a novel local visual grounding reward.

\vspace{1mm}
\noindent\textbf{Global and Format Rewards.}
For fair comparison, we adopt the exact same format reward, score regression reward, and degradation perception reward as Q-Insight~\cite{li2025q}, detailed as follows:
\begin{itemize}
    \item \textbf{Format Reward:} Ensures the response strictly adheres to the structured Chain-of-Thought format. The reward $r_{\text{fmt}}^{(i)} = 1$ if the reasoning process is properly enclosed within \texttt{<think>} tags and the final output is a valid JSON object within \texttt{<answer>} tags; otherwise, $r_{\text{fmt}}^{(i)} = 0$.
    \item \textbf{Score Regression Reward:} For global scoring, we employ a binary reward based on absolute error. Let $scr_{\text{pred}}^{(i)}$ be the predicted score and $scr_{\text{gt}}$ be the ground-truth Mean Opinion Score (MOS). The reward $r_{\text{scr}}^{(i)} = 1$ if $|scr_{\text{pred}}^{(i)} - scr_{\text{gt}}| < \epsilon$, where $\epsilon$ is a predefined acceptable margin; otherwise, $0$.
    \item \textbf{Degradation Perception Reward:} For global degradation perception, we assign $r_{\text{deg}}^{(i)} = 1$ if the predicted distortion category matches the ground truth. Furthermore, an intensity reward $r_{\text{lev}}^{(i)} = 1$ is granted only if both the distortion category and its severity level are correctly predicted.
\end{itemize}

\vspace{1mm}
\noindent\textbf{Local Quality Assessment Rewards.}
Further details regarding the overall local reward structure are provided in Sec. 3.3 of the main paper. Here, we exclusively elaborate on the specific matching process and the calculation of the spatial metrics.

Let $\mathcal{B}_p = \{b_{p,1}, b_{p,2}, \dots, b_{p,m}\}$ denote the set of $m$ predicted bounding boxes, and $\mathcal{B}_{gt} = \{b_{g,1}, b_{g,2}, \dots, b_{g,n}\}$ denote the set of $n$ ground-truth bounding boxes, where each box $b \in \mathbb{R}^4$. To evaluate the spatial localization accuracy, we formulate the bounding box evaluation as an optimal bipartite matching problem. We construct a cost matrix $\mathcal{C} \in \mathbb{R}^{m \times n}$ based on the negative Distance Intersection over Union (DIoU) between every pair of predicted and ground-truth boxes:
\begin{equation}
    \mathcal{C}_{i,j} = -\text{DIoU}(b_{p,i}, b_{g,j}), \quad \forall i \in \{1, \dots, m\}, j \in \{1, \dots, n\}.
\end{equation}

We then apply the Hungarian algorithm to this cost matrix to obtain the optimal bipartite matching $\pi$, which assigns each predicted box to at most one ground-truth box. For the matched pairs $(i, \pi(i))$, we evaluate the matching quality using a predefined DIoU threshold $\tau$ (set to $0.5$ in our experiments). A matched prediction is designated as a True Positive ($TP$) if its DIoU with the corresponding ground-truth box is greater than or equal to $\tau$:
\begin{equation}
    TP = \sum_{i \in \text{matched}} \mathbb{I}(\text{DIoU}(b_{p,i}, b_{g,\pi(i)}) \ge \tau),
\end{equation}
where $\mathbb{I}(\cdot)$ is the indicator function. Consequently, the False Positives ($FP$) and False Negatives ($FN$) are strictly determined by the total number of predictions $m$ and ground truths $n$:
\begin{equation}
    FP = m - TP, \quad FN = n - TP.
\end{equation}

Based on these definitions, the Precision ($\mathcal{P}$) and Recall ($\mathcal{R}$) for the spatial grounding are calculated as:
\begin{equation}
    \mathcal{P} = \frac{TP}{m}, \quad \mathcal{R} = \frac{TP}{n}.
\end{equation}
(Note: If $m=0$, $\mathcal{P}$ is defined as $0$; if $n=0$, $\mathcal{R}$ is defined as $0$). Finally, the discrete spatial matching metric, the $F1$ Score, is computed as the harmonic mean of Precision and Recall:
\begin{equation}
    F1 = \begin{cases}
    \frac{2 \cdot \mathcal{P} \cdot \mathcal{R}}{\mathcal{P} + \mathcal{R}}, & \text{if } \mathcal{P} + \mathcal{R} > 0, \\
    0, & \text{otherwise}.
    \end{cases}
\end{equation}
This rigorous $F1$ Score metric comprehensively evaluates both the precise localization and the quantity alignment of the predicted local artifacts, providing dense and accurate reinforcement signals during the multi-task optimization.
\section{Construction Details of the VGSR-2K Dataset}
\label{sec:dataset_details}

\subsection{Data Composition and Source Distributions}
\label{subsec:data_composition}

The proposed VGSR-2K dataset is meticulously constructed upon two foundational datasets: SRBench~\cite{diffiqa} and DiffIQA~\cite{diffiqa}. Specifically, SRBench~\cite{diffiqa} provides a comprehensive collection of diverse natural scenes enhanced by various super-resolution (SR) architectures, encompassing regression-based, GAN-based, and diffusion-based models, thereby offering a rich spectrum of degradation patterns. Conversely, DiffIQA~\cite{diffiqa} is explicitly designed to address the complex artifacts introduced by modern generative enhancement methods, such as hallucinated textures and structural distortions, operating under a relaxed perfect reference quality assumption. 

To construct our dataset, we employ distinct sampling strategies for these two sources. For the DiffIQA~\cite{diffiqa} subset, we randomly sample 798 high-resolution (HR) images in total. Among these, 698 images are processed through the original synthetic degradation and random enhancement pipeline to formulate 698 $[I_{LR}, I_{SR}]$ data pairs. The remaining 120 HR images undergo exclusively the degradation process to generate 120 pristine $[I_{LR}, I_{HR}]$ data pairs. Crucially, the explicit introduction of these $[I_{LR}, I_{HR}]$ pairs is designed for balancing positive and negative samples, which effectively prevents the policy model from exploiting trivial over-generation loopholes during reinforcement learning. 

For the final detailed data ratios and distribution, please refer to Sec. 4.3 of the main paper.

\subsection{Qualitative Examples of Candidate Generation}
\label{subsec:candidate_generation}

\begin{figure*}[ht]
    \centering
    \includegraphics[width=\linewidth]{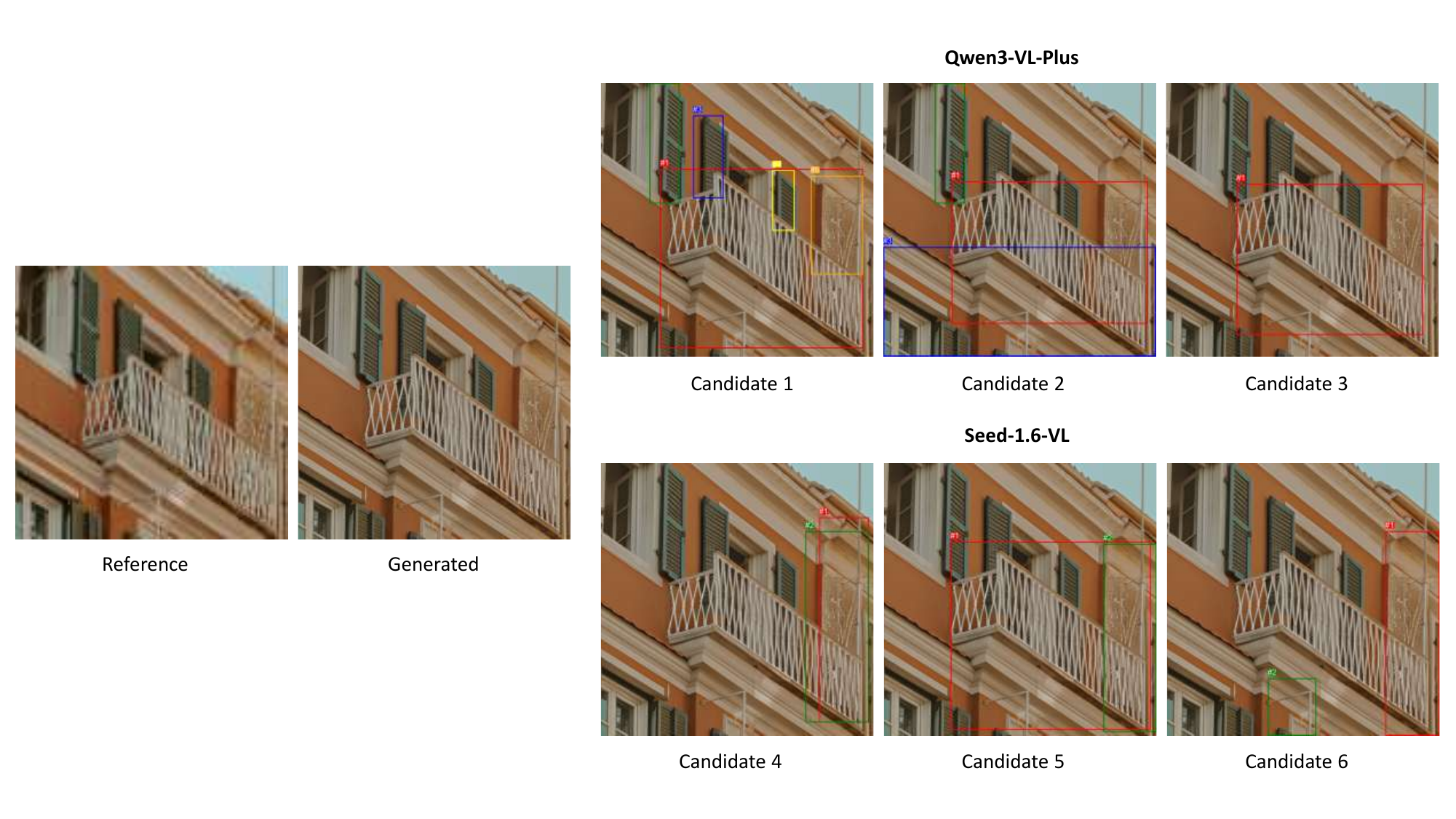}
    \caption{\textbf{Qualitative illustration of the candidate generation process.} This figure presents a given $[I_{Reference}, I_{Generated}]$ image pair alongside its corresponding six different candidate generation results. By providing such a diverse set of candidate bounding boxes, human annotators are empowered to conduct more comprehensive screening and rigorous cross-correction in the subsequent annotation stages.}
    \label{fig:candidate_generation}
\end{figure*}

As detailed in Sec. 4.2 of the main paper, the spatial ground truths of the VGSR-2K dataset are curated through a rigorous four-stage human-in-the-loop annotation pipeline. This subsection aims to provide qualitative visual examples, as illustrated in Fig.~\ref{fig:candidate_generation}, to explicitly demonstrate the output diversity of Stage 2 (MLLM-driven candidate generation). By leveraging two multi-modal foundation models (Qwen3-VL-Plus~\cite{qwen3} and Seed-1.6-VL~\cite{seed1.5}) to generate multiple distinct proposals, this stage ensures a high degree of spatial diversity, thereby supplying rich and comprehensive candidate material for the subsequent rigorous human screening and cross-correction.

\section{Additional Experimental Setup}
\label{sec:experimental_setup}

In this section, we provide extended details regarding the experimental setups, including the evaluation benchmarks, the selection of baseline models, and the specific implementation configurations for our multi-task training framework. 

\subsection{Evaluation Datasets}
\label{subsec:evaluation_datasets}

For the score regression task, we evaluate our model on diverse IQA benchmarks, which are categorized into four groups: (a) in-the-wild datasets including KonIQ~\cite{koniq}, SPAQ~\cite{spaq} and LIVE-Wild~\cite{live}; (b) synthetic distortion datasets (KADID~\cite{lin2019kadid}, CSIQ~\cite{csiq}); (c) model-processed distortions (PIPAL~\cite{pipal}); and (d) AI-generated images (AGIQA~\cite{agiqa}). Following Q-Insight~\cite{li2025q}, we partition the KonIQ~\cite{koniq} dataset, utilizing approximately $7,000$ images for training, with the MOS normalized to the range of $[1, 5]$. The remaining datasets are exclusively used to evaluate the out-of-distribution (OOD) generalization capability of the models. 

For the degradation perception task, we utilize the DQ-495K~\cite{dq} dataset. Specifically, we randomly sample $7,000$ images containing single distortions for training. Crucially, since the test set of Q-Insight~\cite{li2025q} remains unreleased, we establish a fair evaluation benchmark by randomly sampling an additional $1,000$ non-overlapping and identically distributed images from DQ-495K to serve as our test set.

\subsection{Baseline Models}
\label{subsec:baselines}

To thoroughly validate the effectiveness and comprehensiveness of our VGIQA framework on score regression task, we compare VGIQA with handcrafted methods including NIQE~\cite{niqa} and BRISQUE~\cite{bri}; non-MLLM deep-learning methods including NIMA~\cite{nima}, MUSIQ~\cite{ke2021musiq}, CLIP-IQA+~\cite{wang2023exploring}, ManIQA~\cite{yang2022maniqa}; and recent MLLM-based methods such as C2Score~\cite{Compare2Score}, Q-Align~\cite{wu2023q}, DeQA-Score~\cite{deqa}, Q-Insight~\cite{li2025q}, and a supervised fine-tuned Qwen~\cite{qwen2.5}. For a fair comparison, all methods (except handcrafted ones) are trained on the KonIQ~\cite{koniq} dataset, and all MLLM-based methods utilize approximately 7B parameters.

\subsection{Implementation Details}
\label{subsec:implementation_details}

During the cold-start phase, the foundation model (Qwen2.5-VL-7B-Instruct~\cite{qwen2.5}) is trained for 3 epochs with a batch size of 16. In the subsequent GRPO phase, the model is jointly optimized for 10 epochs. Following Q-Insight~\cite{li2025q}, the generation number $N$ is set to 8, the weight of the KL divergence penalty $\beta$ is set to $1 \times 10^{-3}$, the weights $\alpha_1$ and $\alpha_2$ are set to 0.25 and 0.75, respectively, and the threshold $\epsilon$ is set to 0.35. We employ the AdamW optimizer with an initial learning rate of $1 \times 10^{-6}$ that linearly decays to $1 \times 10^{-9}$. Furthermore, the local reward weighting coefficient $\omega_{\text{local}}$ decays linearly from 5.0 to 1.0 during training. To balance the computational overhead under a constrained memory budget, we adopt an asymmetric batch size setting: 8 for the memory-intensive local quality assessment task, and 16 for both degradation perception and score regression tasks. All training of both phases is conducted on 8 compute nodes (80 GB each).

\section{Designed Prompts}
\label{sec:prompts}

The prompts designed for each task in our framework are detailed in Table~\ref{tab:prompts}. Specifically, the input for the score regression and degradation perception tasks consists of a task-specific prompt and a single evaluated image, with the expected outputs being a continuous quality score, and the distortion class alongside its severity level, respectively. For the local quality assessment task, the input includes a reference-generated image pair and a prompt, outputting precise spatial bounding boxes coupled with natural language reasoning for the localized regions. 

To ensure fair comparisons with existing holistic evaluation methods, our system prompt, score regression prompt, and degradation perception prompt strictly follow the designs of Q-Insight~\cite{li2025q}. Crucially, We introduce a new prompt for local quality assessment. By explicitly incorporating \texttt{<think>} and \texttt{<answer>} tags, this prompt guides the model through a rigorous ``look and then judge'' Chain-of-Thought (CoT) reasoning process. This explicit CoT mechanism effectively enhances both the interpretability of spatial grounding and the accuracy of final quality judgments.

{
\footnotesize
\begin{longtable}{@{}p{0.96\linewidth}@{}}
\caption{\textbf{Complete prompt templates used in our multi-task joint training framework.} The system prompt, degradation perception prompt, and score regression prompt are directly adopted from Q-Insight~\cite{li2025q}, while the local quality assessment prompt is our newly designed core contribution.}
\label{tab:prompts} \\

\toprule
\endfirsthead

\multicolumn{1}{c}%
{{\bfseries Table \thetable\ -- continued from previous page}} \\
\toprule
\endhead

\bottomrule
\endfoot

\bottomrule
\endlastfoot

\textbf{System Prompt (Shared)} \\
\midrule
A conversation between User and Assistant. The user asks a question, and the Assistant solves it. The assistant first thinks about the reasoning process in the mind and then provides the user with the answer. The reasoning process and answer are enclosed within \texttt{<think>} \texttt{</think>} and \texttt{<answer>} \texttt{</answer>} tags, respectively, i.e., \texttt{<think>} reasoning process here \texttt{</think>}\texttt{<answer>} answer here \texttt{</answer>}. \\

\midrule
\textbf{Prompt for Local Quality Assessment (Ours)} \\
\midrule
You are an expert in super-resolution (SR) image quality assessment. You are provided with: \\
First image: Low-Resolution (LR) reference image (ground truth basis) \\
Second image: Super-Resolved (SR) output image \\[2mm]

Your task is to carefully compare the two images to detect, localize, and evaluate all regions in the SR image where the generation is problematic (e.g., hallucinations, distortions, or objective degradations). \\[2mm]

\textbf{Evaluation Focus:} For every problematic region identified, provide: \\
\textbf{Localization:} Precise bounding box coordinates that tightly enclose the problematic region. \\
\textbf{Reasoning:} A concise natural-language explanation describing why this region is problematic (e.g., hallucinated texture, structural distortion, inconsistent semantics, or excessive blurring). \\[2mm]

\textbf{Important Constraints:} \\
Only report regions that are \textbf{objectively degraded or hallucinated}. Do not flag naturally blurry or ambiguous areas in the LR as "errors" unless the SR image \textbf{exaggerates, fabricates, or distorts} the content significantly. \\[2mm]

\textbf{Process Requirements (Chain-of-Thought):} Use a step-by-step reasoning process detailed within \texttt{<think>} and \texttt{</think>} tags. Your reasoning should follow this flow: \\
\textbf{Observe the LR image:} What structures, textures, or objects are present? What is the baseline level of detail? \\
\textbf{Inspect the SR image:} Look for specific artifacts. Where does it add unrealistic details (hallucination)? Where is it blurry? Where does it contradict the LR content? \\
\textbf{Assess per region:} For each identified issue, describe the type of degradation and justify why it constitutes a problem. \\
\textbf{Locate \& Describe:} Determine the bounding box coordinates and write a concise reason for each issue. \\[2mm]

\textbf{Output Format:} After your reasoning, in the \texttt{<answer>} tag, output a \textbf{strict JSON object} with a top-level key "issues" containing a list of issue dictionaries. Each dictionary must include: \\
"bbox\_2d": The bounding box coordinates [x1, y1, x2, y2] of the issue. \\
"reason": A concise natural-language explanation of why this region is problematic. \\
\textbf{If the SR image contains no obvious problems or degradations compared to the LR image, the "issues" list must be empty (i.e., "issues": []).} \\

\midrule
\textbf{Prompt for Degradation Perception} \\
\midrule
Analyze the given image and determine if it contains any of the following distortions: "noise", "compression", "blur", or "darken". If a distortion is present, classify its severity as "slight", "moderate", "obvious", "serious", or "catastrophic". Return the result in JSON format with the following keys: "distortion\_class": The detected distortion (or "null" if none). and "severity": The severity level (or "null" if none). \\

\midrule
\textbf{Prompt for Score Regression} \\
\midrule
What is your overall rating on the quality of this picture? The rating should be a float between 1 and 5, rounded to two decimal places, with 1 representing very poor quality and 5 representing excellent quality. Return the final answer in JSON format with the following keys: "rating": The score. \\
\end{longtable}
}

\end{document}